\documentclass[twocolumn]{article}

\usepackage[utf8]{inputenc}
\usepackage{amsfonts,amssymb,amsmath}
\usepackage{graphicx}
\usepackage[authoryear]{natbib}
\usepackage{url}

\usepackage{epstopdf}
\usepackage{verbatim}
\usepackage{mathrsfs}
\usepackage{booktabs}

\title{An open dataset of neural networks for hypernetwork research}

\author{David Kurtenbach, Lior Shamir \\ Kansas State University \\ Department of Computer Science \\ Manhattan, KS 66506}   % Chibuike Bze
\date{}

\begin{document}

\maketitle

\abstract{
Despite the transformative potential of AI, the concept of neural networks that can produce other neural networks by generating model weights (hypernetworks) has been largely understudied. One of the possible reasons is the lack of available research resources that can be used for the purpose of hypernetwork research. Here we describe a dataset of neural networks, designed for the purpose of hypernetworks research. The dataset includes $10^4$ LeNet-5 neural networks trained for binary image classification separated into 10 classes, such that each class contains 1,000 different neural networks that can identify a certain ImageNette V2 class from all other classes. A computing cluster of over $10^4$ cores was used to generate the dataset. Basic classification results show that the neural networks can be classified with accuracy of 72.0\%, indicating that the differences between the neural networks can be identified by supervised machine learning algorithms. The ultimate purpose of the dataset is to enable hypernetworks research. The dataset and the code that generates it are open and accessible to the public. % at \url{https://github.com/davidkurtenb/Hypernetworks_NNweights_TrainingDataset}.
}

\section{Introduction}
\label{introduction}

The concept of hypernetworks \citep{ha2016hypernetworks,krueger2017bayesian,zhang2018graph,chauhan2024brief,knyazev2021parameter,schurholt2022hyper,yoo2024hyper} is used to describe a higher level model that is capable of producing a separate neural network by generating model weights. It is a type of meta-learning architecture where the hypernetwork produces weights for a new network or target network. % With the growing applications of generative artificial intelligence, the maturity of GANs, Transformers, and Encoder/Decoder architectures are expanding in terms of applications. 

Implementing a hypernetwork using neural networks would therefore require neural networks that can perform tasks they were not necessarily trained on. Hypernetworks are capable of generating entirely new models that do not require training, and are not informed through any type of transfer learning or one/few shot learning approach. This shifts neural networks from their initial design, and requires new training and inference techniques that can satisfy the challenging needs of hypernetworks.

Although several approaches have been proposed to address hypernetworks, this area of advanced computing is still generally considered somewhat underexplored. Approaching hypernetworks from a generative framework requires that the training data be given special considerations provided the complexity of the task. While hypernetwork research has explored a variety of approaches, hypernetwork training data is commonly based on conditioning input such as task embeddings, feature distributions, or latent variables \cite{ha2016hypernetworks}. The training data presents common challenges such as high dimensionality and overall generalization.

Despite the efforts, the development of hypernetworks is a challenging task, and research efforts are still being continued. Here we prepared the first dataset of neural networks designed for hypernetwork research. The ultimate purpose of the dataset is to provide a model that can generate neural networks rather than training them. 

Datasets of neural networks have been studied in the past \citep{eilertsen2020classifying}. A dataset of neural networks can be used to train a classifier to identify the machine learning problem that it solves. For instance, a neural network can classify between neural networks trained on MNIST and neural networks trained on CIFAR \citep{eilertsen2020classifying}. {Other studies aimed at predicting the performance of a neural network classifier \citep{unterthiner2020predicting}. Some} architectures have also been proposed for analyzing weight spaces of neural networks \citep{schurholt2021self,schurholt2022model,navon2023equivariant}. 

While these are based on datasets of neural networks, they were not designed for the purpose of hypernetworks. For instance, the ability to distinguish between a classifier that was trained on MNIST data and a classifier that was trained on CIFAR data does not necessarily provide tools that can be used to generate a classifier in the context of hypernetworks. Therefore, the dataset of neural networks described here is based on a single image dataset, which is {\it Imagenette}. Each class in the dataset contains neural networks trained to identify a certain {\it Imagenette} class. That is done by conceptualizing the problem as a binary classification problem, such that one of the classes contains images from the class of interest, while the other class is a collection of random images from all other classes. Such dataset can be used to support Generative Adversarial Networks \citep{goodfellow2014generative,goodfellow2020generative} that instead of generating text or images can ultimately generate neural networks.

A unique trait of hypernetworks is creating efficiency in the training process when compared to traditional methods of feed forward and back propagation cycles. Primary networks that are lighter and contain a smaller number of parameters can produce larger networks containing a higher number of parameters. 

The ability to generate neural networks can ideally lead to solutions of AI tasks without the need to train a neural network for each specific task. Since the training of a neural network is often computationally demanding, generating neural networks can provide a faster and more energy-efficient solution to the training of neural networks. Additionally, it can also lead to a more general AI system that does not require the collection of large training sets for each specific task. 

The codebase and dataset are available publicly at \url{https://github.com/davidkurtenb/Hypernetworks_NNweights_TrainingDataset} and \url{https://huggingface.co/datasets/dk4120/neural_network_parameter_dataset_lenet5_binary/tree/main}, respectively. Historically, machine learning research has been driven by the availability of benchmark datasets such as {\it ImageNet} \citep{deng2009imagenet}, among many others \citep{samaria1994face,phillips1998feret,lecun1998gradient,klimt2004enron,shamir2008iicbu,krizhevsky2009learning,mcfee2012million,dueben2022challenges,cohen2017emnist,moscato2021benchmark,wu2018moleculenet,khan2014painting,lin2014microsoft,sinka2002large,eze2024analysis,thiyagalingam2022scientific,tschalzev2025unreflected} that enabled the advancement of the field. 

These benchmark datasets served as substantial factors in the rapid progression of machine learning and artificial intelligence. They provide researchers with convenient access to data, allowing researchers to focus on the development of their algorithms. As benchmarks, they also allow researchers to compare the performance of their algorithms by applying different algorithms developed by different research teams to the same datasets. For instance, the sub-field of automatic face recognition was powered by the availability of face datasets such as ORL \citep{samaria1994face} or FERET \citep{phillips1998feret}. Similarly, the task of automatic object recognition benefited substantially from benchmark dataset such as ImageNet \citep{deng2009imagenet} among many others. Since benchmark datasets of neural networks are not yet available, the availability of the open dataset can assist in the advancement of hypernetworks research.

\section{Background}
\label{Background}

% The concept of hypernetworks is a meta-learning paradigm that has been researched for its applications related to continual learning through the generation of target network model weights \citep{ha2016hypernetworks,krueger2017bayesian,chauhan2024brief}. 

While there are multiple research efforts around the study of hypernetworks and their applications, the subfield is somewhat nascent, with ample areas to be further explored. The core idea of leveraging a higher-order neural network that sometimes contains smaller number of parameters than the target model to generate weights of a separate neural network is a concept that shifts from the ``typical" manner neural network are created. The concept gained its initial traction with the work of \cite{ha2016hypernetworks}, looking for frameworks to expand the existing methods of training a neural network.  

For instance, \citep{schurholt2022hyper} used hyper-representations with layer-wise loss normalization to aggregate knowledge from model zoos. That allowed to generate new models based on that knowledge. 

Bayesian hypernetworks \citep{krueger2017bayesian} provide an expansion of Bayesian deep learning that can transform noise distribution to a distribution with the parameter of a different neural network. It has been demonstrated to be more resistant to adversarial data \citep{krueger2017bayesian}.

% Hypernetworks can leverage methods to generate weights of another, new neural network. This capability has potential to advance continual learning by offering a solution to catastrophic forgetting by creating adaptive weights. 

% A unique trait of hypernetworks is creating efficiency in the training process when compared to traditional methods of feed forward and back propagation cycles. Primary networks that are lighter and contain a smaller number of parameters can produce larger networks containing a higher number of parameters. 

Applications of hypernetworks have seen a number of use cases with variety of applicability. Their potential has spread across multiple domains such as meta-learning, continual learning, neural architecture search, and reinforcement learning \citep{ehret2021continuallearningrecurrentneural, vonoswald2022continuallearninghypernetworks, zhang2020graphhypernetworksneuralarchitecture,huang2021continualmodelbasedreinforcementlearning}. Particularly, they have the ability to train neural networks in cases of limited training data with few-shot learning.

For instance, hypernetworks have been used to improve Continual Learning. By using the concept of task-conditioned hypernetworks, it has been shown that is was possible to overcome the problem of catastrophic forgetting in ``standard" artificial neural networks trained on several different tasks \citep{vonoswald2022continuallearninghypernetworks}. 

The task of Continual Learning using hypernetworks was also studied by \citep{huang2021continualmodelbasedreinforcementlearning}, using task-conditioned hypernetworks to make learning sufficiently fast. The use of these hypernetworks make on-the-fly learning practical, and therefore allowing to avoid the relatively long response time typical to stationary learning models.  

The concept of Graph HyperNetworks was used to identify the most effective neural network architecture for a certain machine learning problem without the computationally challenging need to train and test all of these architectures 
\citep{zhang2020graphhypernetworksneuralarchitecture}.

Hypernetworks have also been found effective in representation of conditional sentences \citep{yoo2024hyper}. That is done by embedding pre-computed conditions into the corresponding layers, allowing the sentence to be handled differently based on the condition.

Hypernetworks have demonstrated a theoretical value in their application to advance continual learning by resolving catastrophic forgetting. Where traditional neural networks have adjusted model weights during the training process, those weights are then static until the model is retrained. Hypernetworks redefine that paradigm by proposing the notion of dynamic weights. The application of a dynamic weight schema serves as a manner to improve network adaptability and performance \citep{ha2016hypernetworks}.

While the study of hypernetworks presents promising potential, they are not without their challenges. Hypernetworks have faced stability and scalability concerns where the models grow increasingly complex  \cite{keynan2021recomposingreinforcementlearningbuilding}. These challenges are only amplified with computational requirements that have also been difficult to overcome. The relationship between a hypernetwork and target network must be carefully designed in their architecture. Another significant challenge of neural networks is having access to a robust and relevant training dataset. The work covered in this paper aims to begin resolving this challenge, and creating a path toward novel applications of hypernetworks in conjunction with generative approaches.

\section{A dataset of neural networks}
\label{dataset}

The dataset contains 10$^4$ instances of neural networks, divided into a total of 10 classes. Each neural network is a two-way, one-versus-all image classifier, and each class contains 1,000 neural networks that can identify the images of that class. The different classes are taken from the {\it Imagenette} dataset \citep{deng2009imagenet}, specifically the Imagenette 320px V2 dataset with classes 0: Tench, 1: English Springer, 2: Cassette Player, 3: Chain Saw, 4: Church, 5:French Horn, 6: Garbage Truck, 7: Gas Pump, 8: Golf Ball, and 9: Parachute. 

{\it Imagenette} is a well-studied benchmark dataset in a mature stage in its life-cycle. That allows to minimize risks such as missing data, imbalanced classes, or label accuracy, which can be a problem with new datasets \citep{gong2023survey}.  

The code repository also includes model performance metrics, aggregated by class and performance plots for each of the 10$^4$ models. Additionally, to further drive accessibility, the model parameters of the 10,000 LeNet-5 binary classifiers have been compiled into two files. One file condenses the weights and biases by model, referred to as modelwise. The individual parameters for each model are captured by model, and provided as a single flattened tensor. The other file captures parameters across classes by layer, referred to as layerwise. Each of the 10 classes parameters are saved by layer. For example, the class ``church" and ``conv2d" dictionary contains the parameters (combined weights and biases) for the first convolutional layer for all 10,000 LeNet-5 models trained for binary, one-vs-all classifications of a church.  

To generate a dataset of neural networks, each neural network is trained as a two-way classifier. The network is trained such that all images of the first class are images taken from one class of Imagenette. The images of the other class are taken randomly from all other Imagenette classes. Each model training dataset contains 9-10\% of the target class.

That leads to 10 classes such that each class contains 1,000 neural networks that can identify images of one class from all other classes. The dataset is therefore balanced \citep{sinka2002large}. All models were trained for 25 epochs and achieved average accuracy of 91.5\%. Because the images in the other class are selected randomly, every neural network is different. That leads to a dataset of neural networks such that each class contains a large number of neural networks. Each neural network in the dataset was trained with different images, and therefore it is different from the other neural networks in that class.

The architecture that was used for this dataset is LeNet-5 \citep{lecun1998gradient}. The motivation for selecting a relatively simple architecture was to ensure that the generation of the dataset is computationally practical. Another reason is avoiding the curse of dimensionality by using an architecture with a lower number of weights compared to other common architectures such as {\it ResNet} or {\it VGG}. A deeper architecture would have a higher number of parameters, making it more challenging to use it for the purpose of generating new neural networks due to the higher dimensionality. 

Training a very large number of neural networks is a computationally intensive task. The training required over twenty seven hours of a powerful computing cluster with more than 10,000 cores. The cluster is made of 1,296 cores of Xeon E5-2690, 1,296 cores of Xeon E5-2680, 2,048 cores of Xeon E5-2683, 2,400 cores of Xeon E5-2630, 1,823 cores of Xeon Gold 6130, 2,176 cores of AMD EPYC 7452, and 96 nVidia GeForce GTX 2080 Ti. That makes a large cluster of total of 11,039 cores.   % https://support.beocat.ksu.edu/Docs/Compute_Nodes

Using a deeper architecture with more parameters would have led to a dataset that would be impractical to generate even with a powerful cluster. Additionally, a relatively simple architecture simplifies the analysis and use of the dataset. Such analysis can include training a neural network that can classify neural networks, or generate neural networks automatically.

Table~\ref{accuracy_all} shows the classification accuracy, precision, recall, and F1 score of the neural networks of the different classes. Since each class contains 1,000 neural networks, and each neural network is trained separately using different data, the performance of the neural networks contained in each class is not expected to be identical.

\begin{table*}[t]
    \caption{Performance metrics of classification models whose weights are used to compile the training dataset per each classes.}
    \label{accuracy_all}
    \vskip 0.15in
    \centering
    \resizebox{\textwidth}{!}{  % Resize table to fit within the text width
    \begin{tabular}{|l|ccc|ccc|ccc|ccc|}
    \toprule
    \hline
    \textbf{class} & \multicolumn{3}{c|}{\textbf{accuracy}} & \multicolumn{3}{c|}{\textbf{precision}} & \multicolumn{3}{c|}{\textbf{recall}} & \multicolumn{3}{c}{\textbf{F1}} \\
    \textbf{} & \textbf{min} & \textbf{max} & \textbf{average} & \textbf{min} & \textbf{max} & \textbf{average} & \textbf{min} & \textbf{max} & \textbf{average} & \textbf{min} & \textbf{max} & \textbf{average} \\
    \midrule
    \hline
    tench & 0.932 & 0.949 & 0.942 & 0.663 & 0.872 & 0.769 & 0.478 & 0.672 & 0.587 & 0.604 & 0.713 & 0.665 \\
    english\_springer & 0.894 & 0.920 & 0.911 & 0.473 & 0.779 & 0.619 & 0.134 & 0.496 & 0.315 & 0.219 & 0.522 & 0.412 \\
    cassette\_player & 0.916 & 0.937 & 0.928 & 0.544 & 0.845 & 0.675 & 0.272 & 0.569 & 0.408 & 0.407 & 0.581 & 0.506 \\
    chain\_saw & 0.897 & 0.908 & 0.903 & 0.349 & 0.933 & 0.576 & 0.008 & 0.127 & 0.068 & 0.015 & 0.214 & 0.120 \\
    church & 0.901 & 0.921 & 0.911 & 0.533 & 0.844 & 0.666 & 0.134 & 0.438 & 0.301 & 0.226 & 0.518 & 0.411 \\
    french\_horn & 0.886 & 0.907 & 0.900 & 0.300 & 0.634 & 0.507 & 0.008 & 0.353 & 0.186 & 0.015 & 0.406 & 0.265 \\
    garbage\_truck & 0.892 & 0.927 & 0.917 & 0.464 & 0.846 & 0.645 & 0.193 & 0.584 & 0.395 & 0.303 & 0.565 & 0.484 \\
    gas\_pump & 0.870 & 0.901 & 0.892 & 0.295 & 0.684 & 0.480 & 0.062 & 0.234 & 0.151 & 0.109 & 0.306 & 0.228 \\
    golf\_ball & 0.898 & 0.919 & 0.912 & 0.496 & 0.836 & 0.658 & 0.128 & 0.434 & 0.298 & 0.222 & 0.494 & 0.407 \\
    parachute & 0.920 & 0.944 & 0.937 & 0.583 & 0.875 & 0.773 & 0.313 & 0.674 & 0.532 & 0.448 & 0.685 & 0.626 \\
    \textbf{all\_classes} & \textbf{0.870} & \textbf{0.949} & \textbf{0.915} & \textbf{0.295} & \textbf{0.933} & \textbf{0.637} & \textbf{0.008} & \textbf{0.674} & \textbf{0.324} & \textbf{0.015} & \textbf{0.713} & \textbf{0.412} \\
    \hline
    \bottomrule
    \end{tabular}
    }
\end{table*}

\subsection{LeNet-5 Model Training Specifications}

The proposed dataset of neural networks contains simple neural networks trained through one-versus-all binary classification models. As mentioned in Section~\ref{dataset}, these neural networks follow the LeNet-5 architecture. The total number of trainable parameters for each model is 91,481. For comparison, the number of parameters in the common ResNet-50 architecture is over $2\cdot10^6$. Table~\ref{lenet5} summarizes the LeNet-5 architecture and the number of parameters.

\begin{table}
 \caption{The LeNet-5 architecture and the number of parameters.}
 \label{lenet5}
 \vskip 0.15in
\begin{center}
\begin{tabular}{||c c c||} 
 \hline
 Layer(type) & Output Shape & Param Num\\ [0.8ex] 
 \hline\hline
 Conv2D & (None, 32, 32, 6) & 456 \\ 
 \hline
 Conv2D & (None, 12, 12, 16) & 2416 \\ 
 \hline
 Conv2D & (None, 2, 2, 120) & 48,120 \\ 
 \hline
 Dense & (None, 84) & 40,404 \\ 
  \hline
 Dense & (None, 1) & 85 \\ 
 \hline
\end{tabular}
\end{center}
\end{table}

Each model produced a total of 10 arrays containing alternating model weight and bias information, saved in the format of an hdf5 file. The length of each array varies and ranges in parameters from 1 to 48,000. Using the model weights as a source of training data presents a unique approach to the training of hypernetworks. Because each neural network is trained with different images, the distribution of the weights within each class of model is distinct. % and demonstrates a potential proof to be further leveraged for use with hypernetworks. 
Most of the individual model weights were near-zero numbers.

\begin{figure}
    \centering
    \includegraphics[width=1\linewidth]{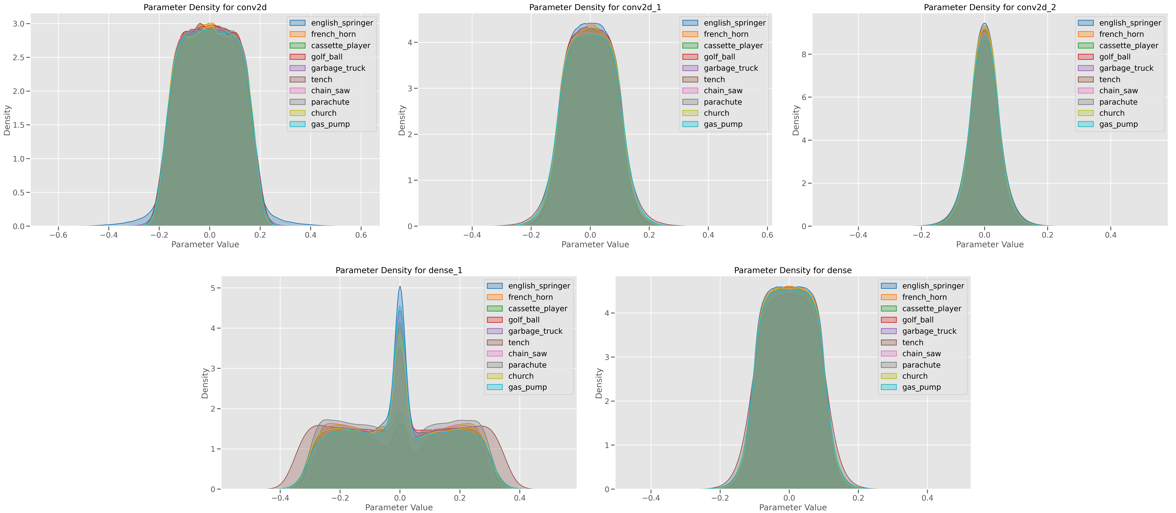}
    \caption{Distribution of class parameters across layers by class.}
    \label{all_class_hist}
\end{figure}

\begin{figure}
    \centering
    \includegraphics[width=1\linewidth]{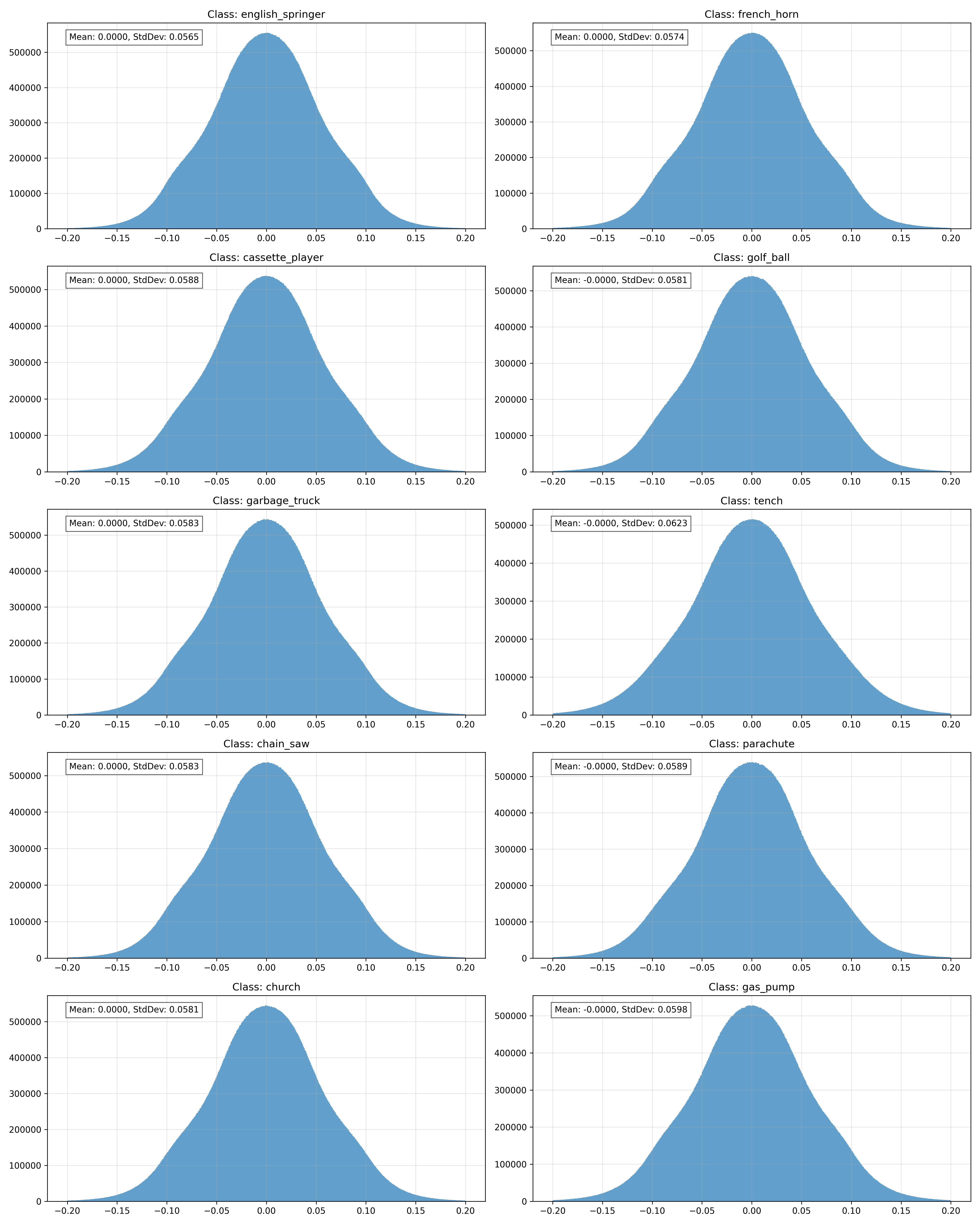}
    \caption{Distribution of class parameters by class. Because each neural network is trained with different images, the parameters are not expected to be identical.}
    \label{byclass_hist}
\end{figure}

Figure~\ref{all_class_hist} displays the distribution of all weights of all classes, and Figure~\ref{byclass_hist} displays the weight distributions separated by class. The plots are scaled to highlight the near-zero distributions of each model due to the large concentration of values within this range. The values of the weights are not identical, which can be expected given that each neural network is trained with a different set of images. The distinct curves for a given class provides evidence of the distinct patterns and features calculated across weight values for the object classification LeNet-5 models. Table~\ref{tab:jsd_values} shows the distribution of common weights in the trained neural networks among the different classes. 

\section{Parameter distribution}

Understanding the distributions and distinction in patterns between layers separated by class is a critical piece in learning characteristics. It is not just the overall distribution by model that is important but should also look at distributional differences of the LeNet-5 model layers. Analysis was performed to better understand the distributions as well as compare divergence between classes.

As parameters traverse the LeNet-5 architecture there is expected reshaping of their distribution. The convolutional approach reduces the total range of distribution, which then undergoes significant transformations as information is passed through the dense layers. Each class has its own unique pattern, but follows a similar profile. The Jensen-Shannon (JS) Divergence was used to assess the level of similarity between class parameter layer distributions. As expected, the parameter distribution at the third and final convolutional layer were the most similar. They demonstrated characteristics that were nearly overlapping when comparing between two different classes. However, this is expected as convolutional layers reduce complexity within the distributions and limit the feature space. This is an aspect of convoloutional layers ability to focus on spatial relationships and employ weight sharing. On the opposite side, the second dense layer was the most diverse layer. This again is expected as the dense layer is connecting all neurons passed by the first dense layer. The effect is to open up the range of the parameter distribution. 

\begin{table}[htbp]
    \centering
    \begin{tabular}{|l|c|c|c|}
        \hline
        \textbf{Layer} & \textbf{Min} & \textbf{Max} & \textbf{Avg} \\
        \hline
        Conv2d & 0.0071 & 0.1001 & 0.0357 \\
        Conv2d\_1 & 0.0050 & 0.0723 & 0.0256 \\
        Conv2d\_2 & 0.0019 & 0.0566 & 0.0200 \\
        Dense & 0.0024 & 0.0680 & 0.0201 \\
        Dense\_1 & 0.0218 & 0.2394 & 0.0904 \\
        \hline
    \end{tabular}
    \caption{Jensen-Shannon Divergence values across different layers}
    \label{tab:jsd_values}
\end{table}

\section{Automatic classification of neural networks}
\label{classification}

To further explore the potential of the dataset in developing hypernetworks, the model weights were used in classification tasks. In demonstrating the ability for the training set to be effectively classified using traditional machine learning and deep learning approaches, one can reason that the training data has ample features within the model weights. This is a primary requirement that leads to the potential of developing hypernetworks with a robust training dataset.

As mentioned in Section~\ref{dataset}, the dataset is fully balanced and contains no missing values. Therefore, classification accuracy higher than mere chance reflects the ability of the classifier to identify between the neural networks. The effectiveness of the classifier was measured by the classification accuracy \citep{sinka2002large}, as well as the specificity, sensitivity, and f1.

\subsection{Classification methods}

Traditional methods of classification were applied to baseline the model performance. Given the high dimensionality of the data, a deep learning model was also applied. Classification was completed with using the layer weights and biases with a total of 91,481 parameters per model. Following standard practices \citep{singh2021impact}, the experiments were performed such that 70\% of the samples were allocated for training, and the rest of the data was used for testing/validation. 

The deep neural network that was used is a fully connected multi-layer perceptron, with three hidden layers of sizes of 256, 128, and 64, with batch normalization. The activation functions are ReLU, and the dropout rate was set to 0.6.

\subsection{Classification results}

The results for the entire model are summarized in Table~\ref{accuracy}. As the table shows, the classification accuracy is far higher than the expected 10\% mere chance, showing that the neural networks can be differentiated from each other by their weights. Naive Bayes achieves the highest classification accuracy of 72\% (p$<10^{-5}$).

\begin{table}[h]
\caption{Classification accuracy results of the neural network dataset applied to full model parameters. The results show that machine learning algorithms can analyze a neural network and identify what the neural network is trained to classify.}
\label{accuracy}
\footnotesize
\vskip 0.11in
\begin{center}
\setlength{\tabcolsep}{3pt}
\begin{tabular}{|l c c c c|} 
 \hline
 Model & Accuracy & Precision & Recall & F1 \\ [0.8ex] 
 \hline\hline
 Random Forest & 0.49 & 0.47 & 0.49 & 0.46 \\ 
 \hline
 Support Vector Machine & 0.25 & 0.22 & 0.25 & 0.22\\
 \hline
 Naive Bayes & 0.72 & 0.73 & 0.72 & 0.72 \\
 \hline
 XGBoost & 0.69 & 0.69 & 0.68 & 0.68 \\ 
 \hline
 Logistic Regression & 0.10 & 0.10 & 0.10 & 0.10 \\
 \hline
 DNN & 0.19 & 0.13 & 0.19 & 0.13 \\
 \hline
\end{tabular}
\end{center}
\end{table}

\begin{figure*} %[!tbp]
  \centering
%  \begin{minipage}[b]{0.4\textwidth}
    \includegraphics[width=\textwidth]{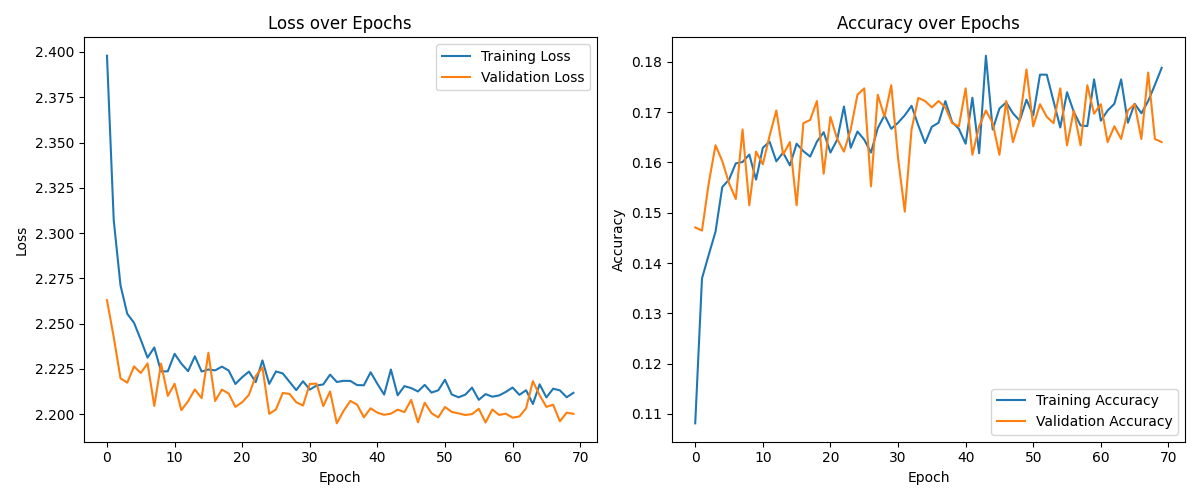}
    \caption{Loss and accuracy curves of deep learning model with all weights.}
    \label{loss_all_weights}
%  \end{minipage}
\end{figure*}

The results observed using deep learning classification capture some of the challenges within the subfield of hypernetworks. The high dimensionality of model weights is challenging to work with and prone to overfitting. Even within this example, practices such as batch normalization, dropout, regularization, random search parameter tuning, and experimentation with model architecture were used with minimal success in terms of improving accuracy. Figure~\ref{loss_all_weights} shows the loss and accuracy of the deep learning model when using all weights.

\section{Discussion}
\label{discussion}

The dataset of 10$^4$ neural networks introduced here was designed specifically for hypernetwork research. Therefore, it is important that the neural networks can be distinguishable through an automatic process. That can show that the weights of the different neural networks exhibit different patterns that are identifiable by machine learning algorithms.  

An attempt to use a classifier that can predict the class that a neural network identifies showed that the classifier can identify the class through the weights of the neural network in accuracy far higher than mere chance. That provides an indication that the dataset can be used for studies that involve machine learning.

For the purpose of automatic classification of neural networks, the deep neural network did not perform well compared to other algorithms, while Naive Bayesian networks showed the best performance. Naive Bayes assumes that each parameter is independent, and therefore performs well when the input variables are independent from each other \citep{friedman1997bayesian}. Weights in a neural network are independent values. For instance, weight in neural networks normally cannot be predicted from other weights, unlike other types of data such as values of pixels in an image. It can therefore be expected that the Naive Bayes provides the best classification accuracy for this specific task.    
  
The fact that the neural networks can be separated using machine learning provides an indication of the existence of patterns in the weights. The expected presence of such patterns is also an indication that such distributions can be produced by generative AI for the purpose of hypernetworks. Generative AI if often used to generate images, audio, video, text, and code \citep{li2023assessing}. Tools such as AlphaEvolve \citep{cui2021alphaevolve} show that it can also be used to generate new algorithms. Here we provide research resources for exploring the contention that generative AI can also be used to generate artificial neural networks.

% The purpose of the dataset is not necessarily to develop classifiers of neural networks, but the fact that the neural networks can be classified provides an indication that the classes of neural networks are different from each other. Therefore, the dataset can be used for machine learning analysis related to hypernetworks, including unsupervised machine learning and generative AI. 

For the direct purpose of generative AI, the classifier of neural networks shows that a GAN discriminator is possible. The results can also be used as baseline for future algorithms that can classify between neural networks. Improving the classification accuracy can lead to better discriminators.

\section{Conclusion}

Here we introduced an open dataset for the study of hypernetworks. The generation of the dataset involved substantial computing resources, resulting in $10^4$ neural networks separated into 10 classes based on Imagenette data. The purpose of the dataset is to enable the research of hypernetworks. The dataset is open and available to the public. Using a known dataset such as Imagenette to generate the neural networks will allow to better understand the nature of the content of the dataset, but it can also allow to expand the dataset in the future by training new image classes against the Imagenette images. 

While datasets of neural networks exist, the dataset described here is designed specifically for the purpose of hypernetwork research. For instance, it is based on a single dataset, rather than an attempt to distinguish between neural networks trained with two completely different datasets \citep{eilertsen2020classifying}. It also uses the same neural network architecture, as it does not aim at identifying the ideal architecture for a given classification problem \citep{unterthiner2020predicting}.  

The dataset of $10^4$ neural networks separated into 10 classes is definitely far smaller than the number of classes and images in a dataset such as {\it ImageNet}. Another limitation of the dataset is that it is limited to one CNN architecture. Naturally, large datasets of neural networks, require substantial computing resources to generate each sample, and are far more demanding than just adding an image sample to a ``traditional" dataset. When using a more complex CNN architecture the training can require far more powerful computing resources, and a higher number of parameters. Yet, the dataset can provide research infrastructure for the development of the concept of hypernetworks, an can be used for a variety of purposes that include supervised machine learning, unsupervised machine learning, and generative AI.

The dataset is based on the relatively simple LeNet-5 architecture. It can be trained within reasonable time using a powerful computing cluster. Future benchmarks will include other common architectures such as ResNet, although using more complex architectures with a higher number of parameters will require substantially stronger computing resources. A higher number of parameters will also require more complex hypernetworks that can be trained by these neural networks. That will require stronger computing and longer training not merely to generate the dataset, but also to train the hypernetworks.

Future work will also include the development of GANs that can generate neural networks. %One of the tasks enabled by the dataset is the development of Generative adversarial networks \citep{goodfellow2014generative,goodfellow2020generative}. 
While GANs are often used to generate images or text, they can also be used to generate neural networks. That, however, requires a suitable dataset of neural networks that can allow the training of a GAN that generates neural networks. Such GAN will require modification to the commonly used GAN architectures. The availability of datasets of neural networks as described here can enable the development and testing of such GANs. 

\section*{Acknowledgments}

We would like to thank the three reviewers for the helpful comments. This study was supported in part by NSF grant 2148878.

\section*{Data availability}

The code used in this project is available at \url{https://github.com/davidkurtenb/Hypernetworks_NNweights_TrainingDataset}. The dataset is available at \url{https://huggingface.co/datasets/dk4120/neural_network_parameter_dataset_lenet5_binary/tree/main}.

%\twocolumn % Return to two column mode if needed
%\clearpage

\bibliographystyle{apalike}
\bibliography{main_arxiv}

\clearpage
\section{Appendix}
\appendix

\begin{figure*}[h]
    \makebox[\textwidth][c]{
        \includegraphics[width=0.9\textwidth]{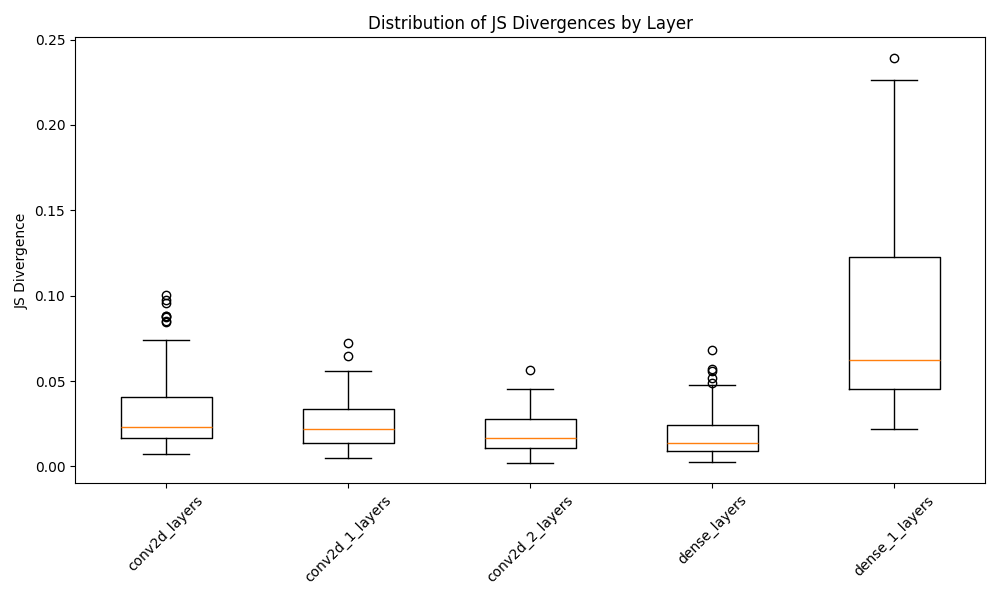}
    }
    \caption{JS divergence distribution by layer.}
    \label{fig:js_divergence_by_layer}
\end{figure*}

\begin{figure*}[h]
    \centering
    \includegraphics[width=0.9\linewidth]{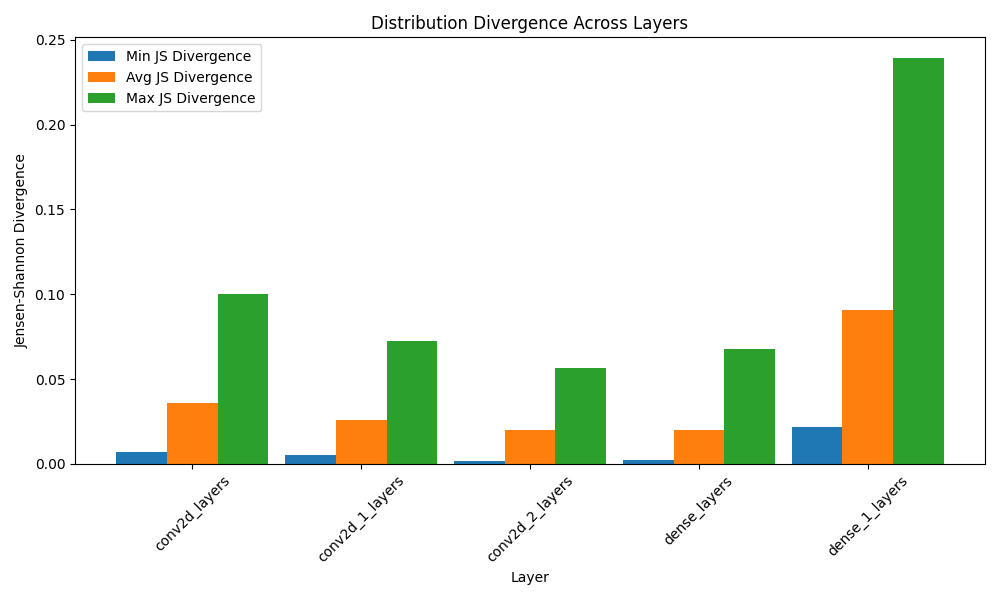}
    \caption{JS diveregence distribution by layer.}
    \label{fig:js_distro_by_layer}
\end{figure*}

\begin{figure*}[t!]
    \makebox[\textwidth][c]{
        \includegraphics[width=0.9\textwidth]{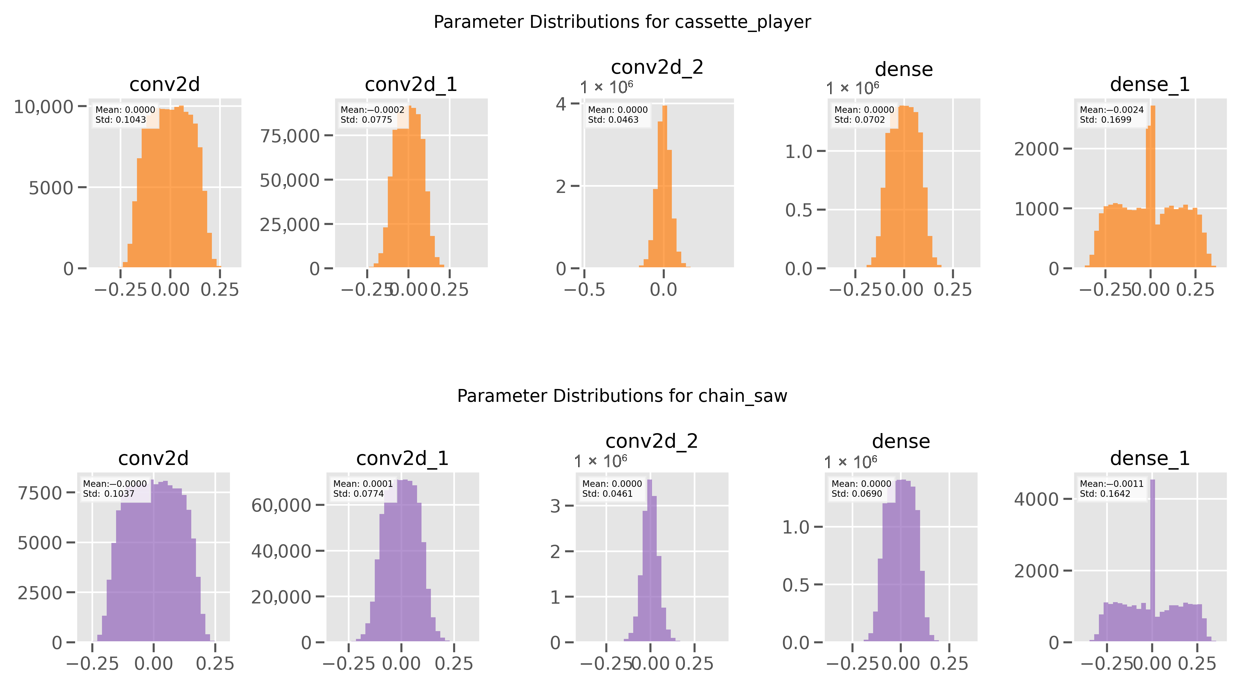}
    }
    \caption{Parameter distribution by layer for classes "cassette\_player" and "chain\_saw".}
    \label{fig:parameter_distribution_by_layer_0}
\end{figure*}

\begin{figure*}[t!]
    \makebox[\textwidth][c]{
        \includegraphics[width=0.9\textwidth]{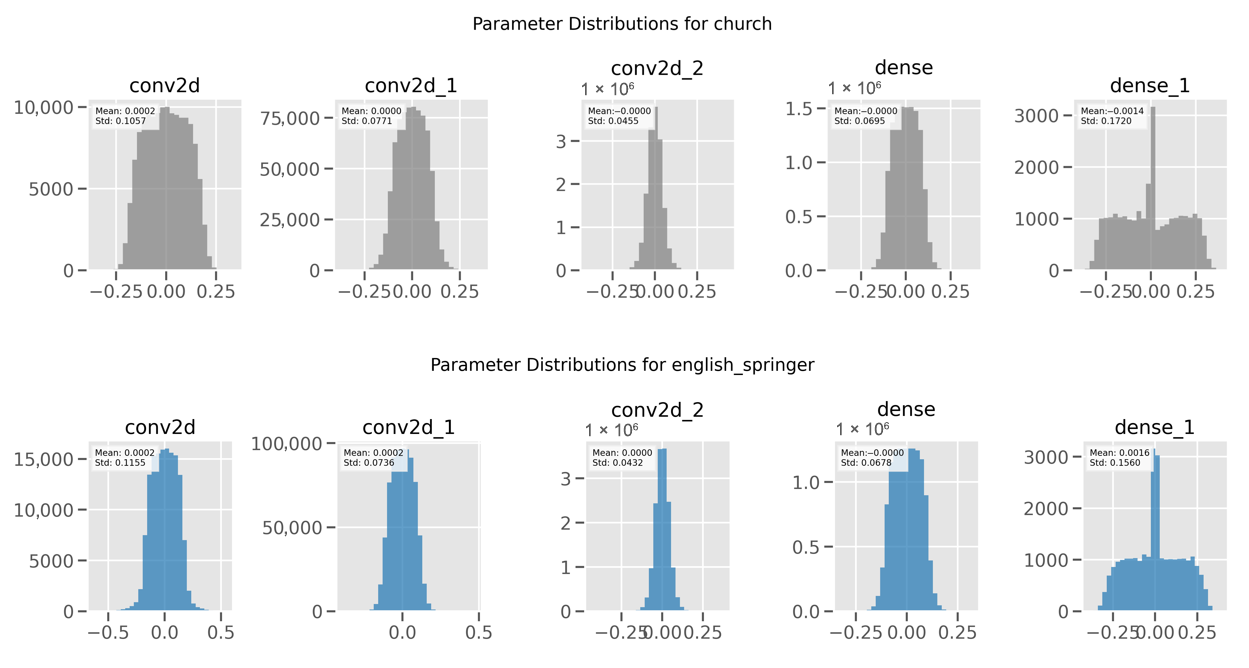}
    }
    \caption{Parameter distribution by layer for classes "church" and "english\_springer".}
    \label{fig:parameter_distribution_by_layer_0}
\end{figure*}

\begin{figure*}[t!]
    \makebox[\textwidth][c]{
        \includegraphics[width=0.9\textwidth]{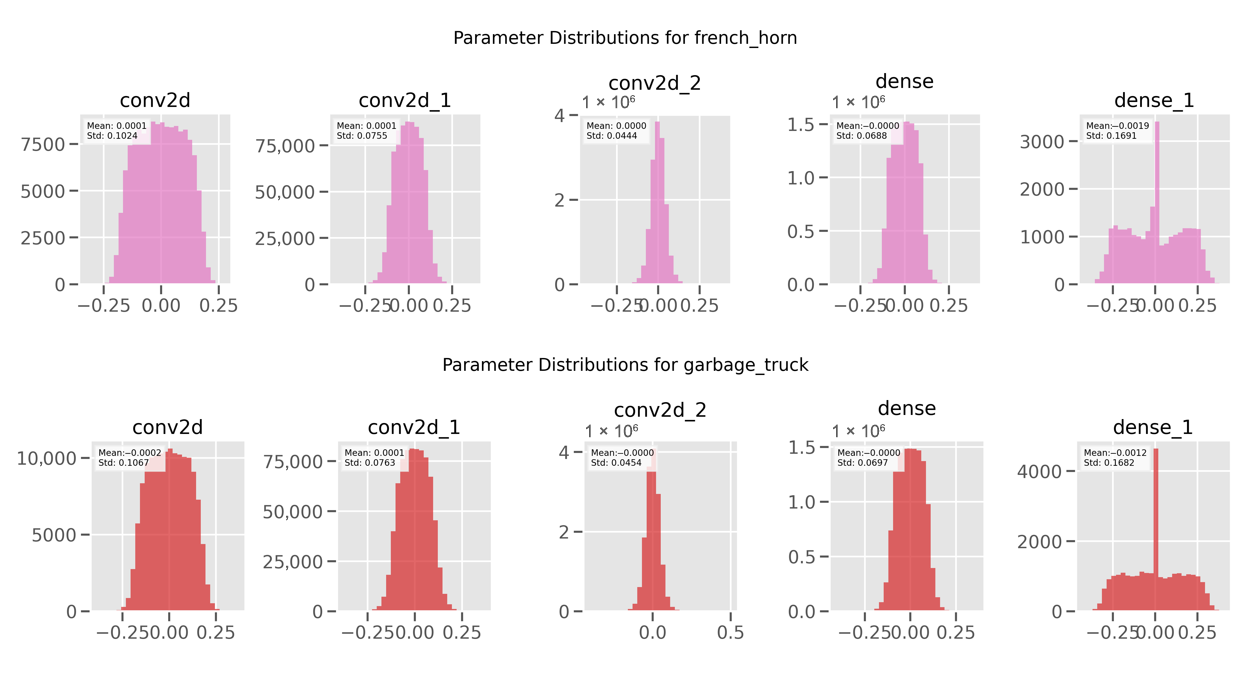}
    }
    \caption{Parameter distribution by layer for classes "french\_horn" and "garbage\_truck".}
    \label{fig:parameter_distribution_by_layer_0}
\end{figure*}

\begin{figure*}[t!]
    \makebox[\textwidth][c]{
        \includegraphics[width=0.9\textwidth]{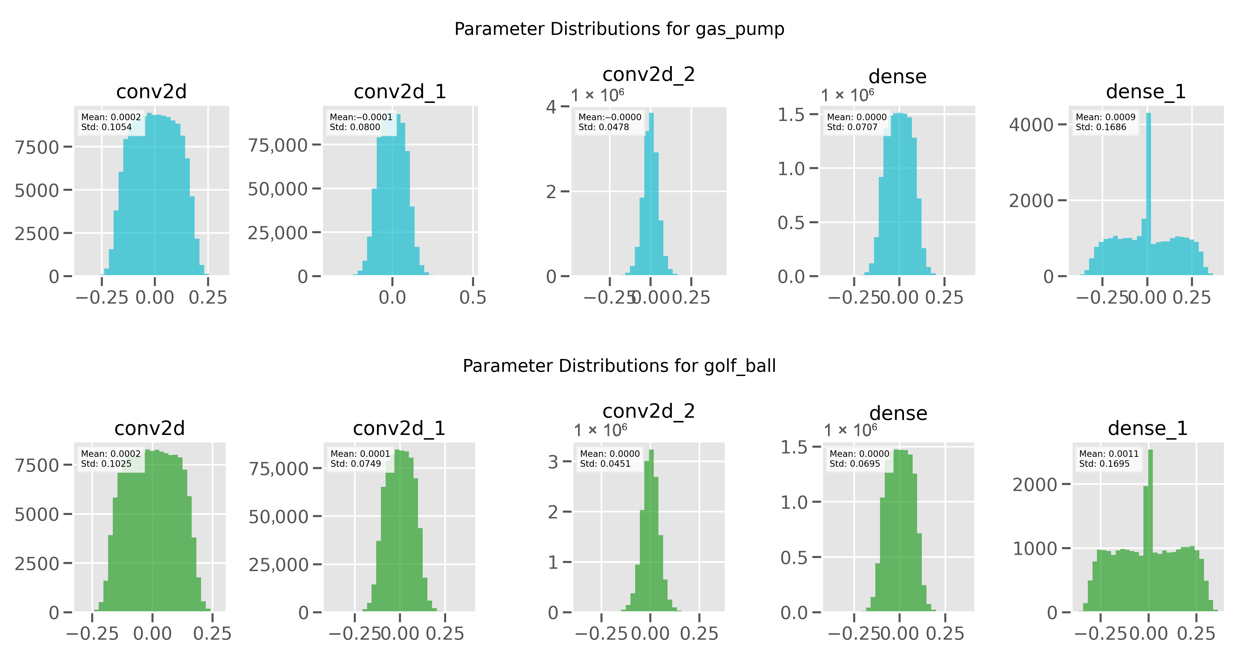}
    }
    \caption{Parameter distribution by layer for classes "gas\_pump" and "golf\_ball".}
    \label{fig:parameter_distribution_by_layer_0}
\end{figure*}

\begin{figure*}[t!]
    \makebox[\textwidth][c]{
        \includegraphics[width=0.9\textwidth]{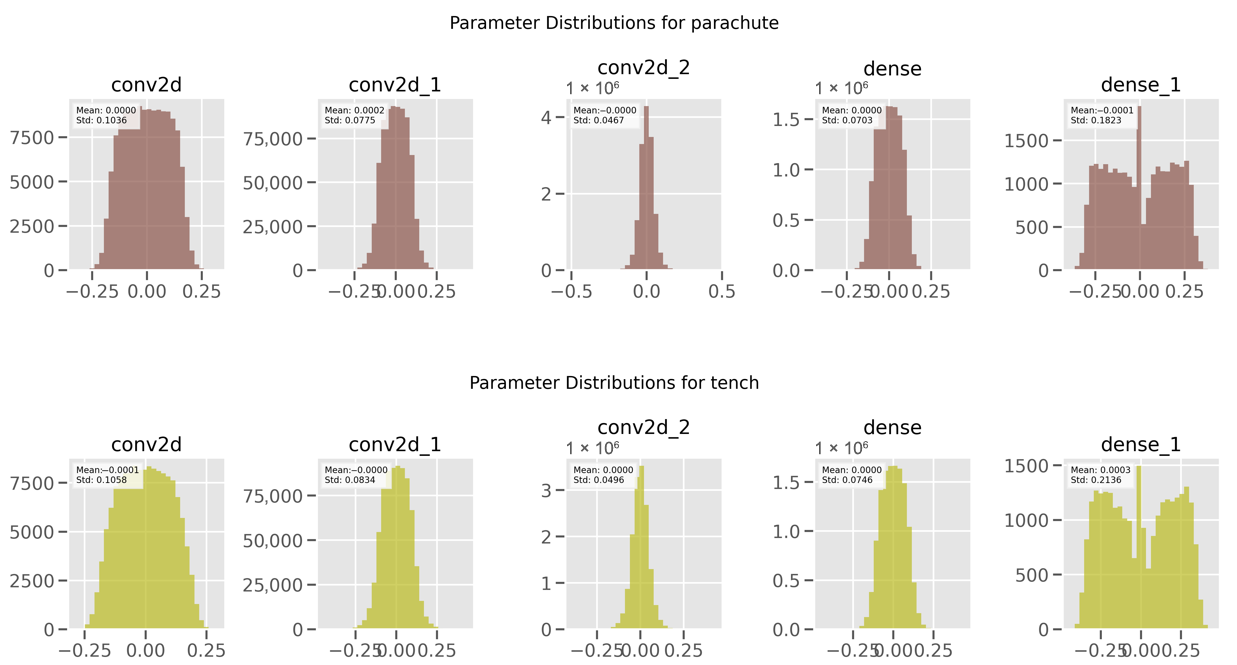}
    }
    \caption{Parameter distribution by layer for classes "parachute" and "tench".}
    \label{fig:parameter_distribution_by_layer_0}
\end{figure*}

\begin{figure*}
    \centering
    \includegraphics[width=1\linewidth]{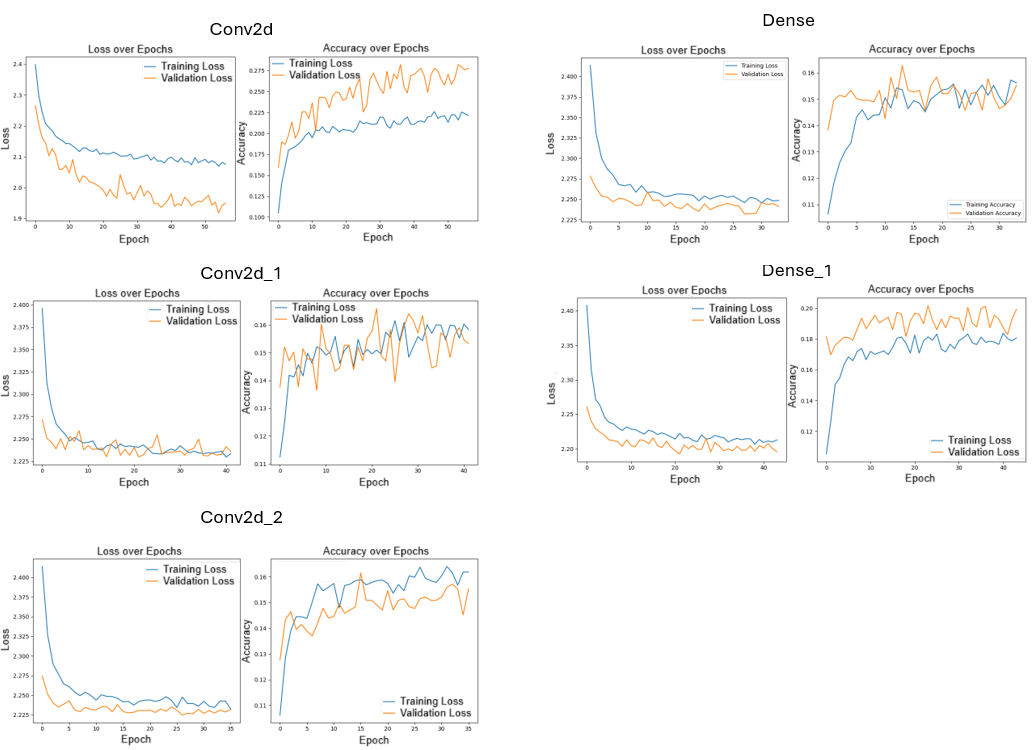}
    \caption{Training/validation loss of DNN classification model by layer.}
    \label{fig:train_val_loss_by_layer_DNN}
\end{figure*}

\end{document}